\documentclass[10pt,twocolumn,letterpaper]{article}

\usepackage{cvpr}
\usepackage{times}
\usepackage{epsfig}
\usepackage{graphicx}
\usepackage{amsmath}
\usepackage{amssymb}

\usepackage{algorithm}
\usepackage{algcompatible}
\usepackage[noend]{algpseudocode}

\usepackage{booktabs}
\usepackage{cuted}
\usepackage{capt-of}

\usepackage[pagebackref=true,breaklinks=true,colorlinks,bookmarks=false]{hyperref}

\cvprfinalcopy % *** Uncomment this line for the final submission

\def\cvprPaperID{2987} % *** Enter the CVPR Paper ID here

\begin{document}

%%%%%%%%% TITLE
% \title{Universal Lighting Model for Face Registration}
\title{ High-fidelity Face Tracking for AR/VR via Deep Lighting Adaptation}

\author{Lele Chen$^{1,2}$  \quad
Chen Cao$^{^1}$ \quad
Fernando De la Torre$^{1}$ \quad 
Jason Saragih$^1$ \quad
Chenliang Xu$^{2}$   \quad
Yaser Sheikh$^1$ \\
$^1$ Facebook Reality Labs \quad $^2$ Univeristy of Rochester\\
% {\tt\small {l.chen, chenliang.xu}@rochester.edu},\quad {}
}

\maketitle
%\thispagestyle{empty}
% %\vspace{-6mm}
\begin{strip}\centering
\includegraphics[width=0.98\textwidth]{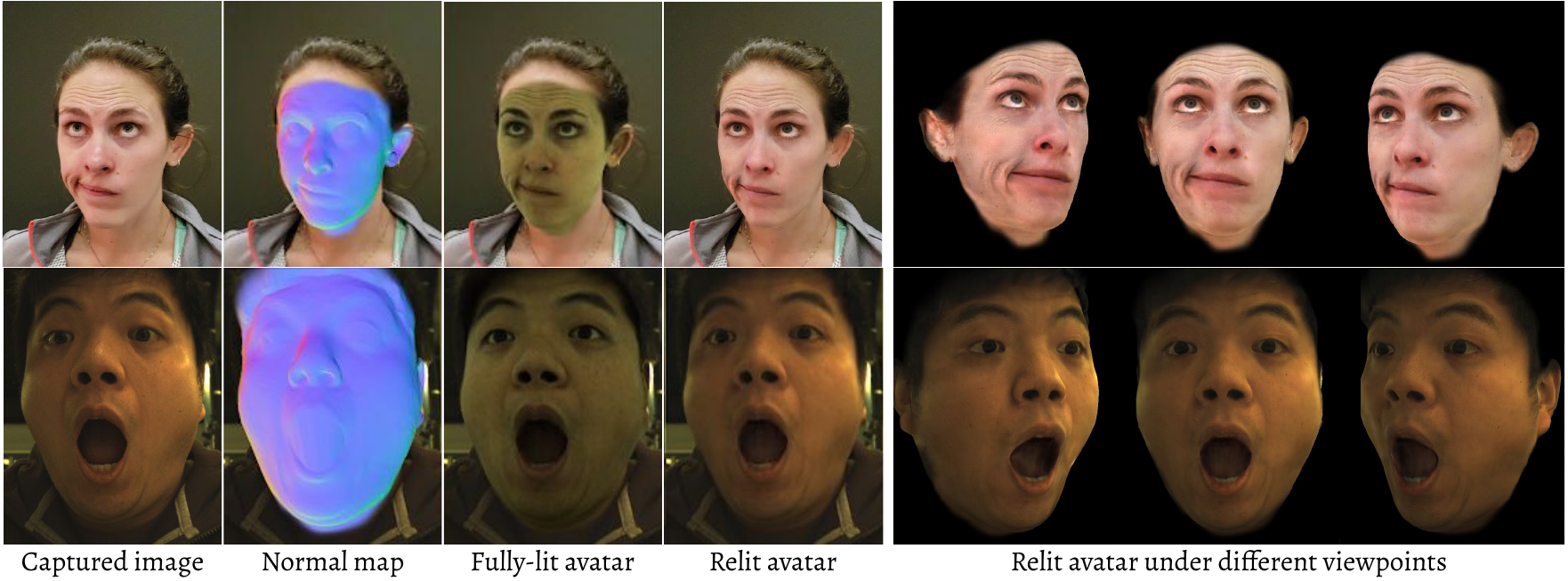}
\captionof{figure}{High-fidelity face tracking results using our method. From left to right: Input image captured by iPhone, normal map, fully-lit avatar (avatar before relighting), relit avatar (avatar after relighting), and relit avatar under different viewpoints. Please notice the specular highlight changes on the avatars under different viewpoints.}
\label{fig:teaser}
\end{strip}
%%%%%% ABSTRACT
\begin{abstract}
% %\vspace*{-4mm} 
3D video avatars can empower virtual communications by providing compression, privacy, entertainment, and a sense of presence in AR/VR. Best 3D photo-realistic AR/VR avatars driven by video, that can minimize uncanny effects, rely on person-specific models. However, existing person-specific photo-realistic 3D models are not robust to lighting, hence their results typically miss subtle facial behaviors and cause artifacts in the avatar. This is a major drawback for the scalability of these models in communication systems (\eg, Messenger, Skype, FaceTime) and AR/VR. This paper addresses previous limitations by learning a deep learning lighting model, that in combination with a high-quality 3D face tracking algorithm, provides a method for subtle and robust facial motion transfer from a regular video to a 3D photo-realistic avatar. Extensive experimental validation and comparisons to other state-of-the-art methods demonstrate the effectiveness of the proposed framework in real-world scenarios with variability in pose, expression, and illumination. Our project page can be found at \href{https://www.cs.rochester.edu/u/lchen63}{https://www.cs.rochester.edu/u/lchen63}. Please visit \href{https://www.youtube.com/watch?v=dtz1LgZR8cc}{https://www.youtube.com/watch?v=dtz1LgZR8cc} for
more visual results.

% Our project page can be found at this \href{https://www.cs.rochester.edu/~cxu22/r/wild-avatar/}{website}. %\url{https://www.cs.rochester.edu/~cxu22/r/wild-avatar/}.

% We show the benefits of our approach in a variety of in-the-wild videos with variability in pose, expression and illumination. 

%While high-fidelity face registration from in-the-wild videos is critical to AR/VR, it is remaining unsolved due to the lack of high-quality tracked 3D mesh and unwrapped texture from data captured by commodity cameras. In this paper, we present a tracking system with an adaptive lighting model to tackle the face registration problem for in-the-wild videos captured under a variety of unknown poses, expressions, and illumination conditions\footnote{Although there is no restriction for lighting sources, our lighting model requires a stable lighting environment in each video.}. Specifically, We learn a lighting model from light-stage data, and then by adapting it to target video and updating the tracking parameters alternatively, our registration system is able to achieve the high-fidelity avatar animation with complex facial motion driven by videos captured by commodity cameras (\eg, iPhone). Through extensive experimental validation and comparisons to other state-of-the-art methods, we demonstrate the effectiveness of the proposed framework in real-world scenarios.
\end{abstract}

%%%%%%%%% BODY TEXT
\section{Introduction}
% Introduce the task.

Currently, video conferencing (\eg, Zoom, Skype, Messenger) is the best 2D available technology for internet communication.  
To allow for more advance levels of communication and sense of presence, Augmented Reality (AR) and Virtual Reality (VR) technologies aim to build 3D personalized avatars, and superimpose virtual objects in the real space. If successful, this new form of face-to-face interaction will allow extended remote work experiences that can improve productivity, reducing cost and stress of commuting, have a huge impact on the environment, and overall improving the work/life balance.

Today most real-time systems for avatars in AR are cartoon-like (\eg, Apple Animoji, Tiktok FaceAnimation, Hyprsense, Loom AI); on the other hand,
digital creators in movies have developed uncanny digital humans using advanced computer graphics technology and person-specific (PS) models (\eg, Siren). While some of these avatars can be driven in real-time from egocentric cameras (\eg, Doug character made by digital domain), building the PS model is an extremely time-consuming and hand-tuned process that prevents democratizing this technology. This paper contributes toward this direction, and it proposes new algorithms to robustly and accurately drive 3D video-realistic avatars from monocular cameras to be consumed by AR/VR displays (see Fig.~\ref{fig:teaser}). 

Model-based photo-realistic 3D face reconstruction/animation from a video has been a core area of research in computer vision and graphics in the last thirty years~\cite{blanz1999morphable,booth2018large,cao2013facewarehouse,cootes2001active,gerig2018morphable, huber2016multiresolution,li2017learning,matthews2004active, paysan20093d, tzimiropoulos2012generic,thies2015realtime,zollhoefer2014,Bee11}. While different versions of morphable models or active appearance models have provided good facial animation results, the existing 3D models do not provide the quality that is needed it for a good immersive viewing experience in AR/VR. In particular, the complex lighting, motion, and other in-the-wild conditions do result in artifacts in the avatar due to poor decouple of rigid and non-rigid motion, as well as, no accurate texture reconstruction. To tackle this problem, we build on recent work on Deep Appearance Model (DAM)~\cite{lombardi2018deep} that learns a person-specific model from a multi-view capture setup. \cite{lombardi2018deep} can render photo-realistic avatars in a VR headset by inferring 3D geometry and view-dependent texture from egocentric cameras . 
%a Plenoptic study, and infer the 3D shape and texture transferring subtle expressions from the head mounted cameras.

This paper extends DAM~\cite{lombardi2018deep} with a new deep lighting adaptation method to recover
subtle facial expressions from monocular videos in-the-wild and transfer them to a 3D video-realistic avatar. The method is able to decouple rigid and non-rigid facial motions, as well as, shape, appearance and lighting from videos in-the-wild. Our method combines a prior lighting model learned in a lab-controlled scenario and adapts it to the in-the-wild video, recovering accurate texture and geometric details in the avatar from images with complex illuminations and challenging poses (\eg profile).  There are two main contributions of our work. First, we provide a framework for fitting a non-linear appearance model (based on a variational auto-encoder) to in-the-wild videos. Second, we propose a new lighting transfer module to learn a global illumination model. Experimental validation shows that our proposed algorithm with deep lighting adaptation outperforms state-of-the-art methods and provides robust solutions in realistic scenarios.

\section{Related Work}
\label{sec:related}

%Indeed, similar problem~\cite{yoon2019self,bai2020deep,liu20193d,chen2019photo,cao2018stabilized,gecer2019ganfit,lin2020towards,zeng2019df2net} has been explored recently. For example, Gecer~\etal~\cite{gecer2019ganfit} train a facial texture generator in UV space with self-supervision as their statistical parametric representation of the facial texture. Their method is able to reconstruct high-quality facial shape and texture in arbitrary recording conditions. Lin~\etal~\cite{lin2020towards} propose a GCN-based network to refine the texture generated by 3DMM based method with facial details from the input image. Zeng~\etal~\cite{zeng2019df2net} decompose the reconstruction into three cascaded stages, where each stage outputs geometry with more details gradually.
% However, they either ignore the lighting or model it with low-order spherical harmonics, which cannot produce hard shadows cast from point light sources, which will decrease the face reconstruction quality in many scenarios. Moreover, the surface information presented by a single image is limited, which leads to some artifacts such as over-smoothing and incorrect expression.

\noindent \textbf {3D Face Tracking.}\quad
3D morphable face models (3DMM)~\cite{blanz1999morphable,booth2018large,cao2013facewarehouse,gerig2018morphable,huber2016multiresolution,li2017learning,paysan20093d} and Active Appearance Models (AAMs)~\cite{cootes2001active,matthews2004active,tzimiropoulos2012generic} have been extensively utilized for learning facial animations from 3D scans and face tracking. These methods produce texture and geometry through the idea of analysis-by-synthesis, where a parametric face model is iteratively adapted until the synthesized face matches the target image. For example, by leveraging the photometric error in both shape and texture, AAMs~\cite{cootes2001active} have shown strong efficiency and expressibility to register faces in images. More recently, Deep Appearance Model (DAM)~\cite{lombardi2018deep} extends the AAMs with deep neural networks in-place of linear generative functions. DAM learns the latent presentation of geometry and texture using a conditional variational autoencoder~\cite{kingma2013auto} and is able to reconstruct a high-fidelity view-dependent avatar with the aid of the multi-view camera system.

\noindent \textbf{In-the-Wild Face Reconstruction.}\quad
Face reconstruction under an in-the-wild scenario is known as a challenging problem since the rigid, lighting, and expression are unknown. For example, the surface information presented by a single image~\cite{guo2020towards,lin2020towards,richardson20163d,sela2017unrestricted,tewari2018self,tewari2017mofa,Zhou2021pose,tran2019learning,yoon2019self,zhu2017face,Zhou_2020_CVPR,Thies19F2F,kim17InverseFaceNet,GZCVVPT16} or even an image collection~\cite{roth2015unconstrained,adaptive-3d-face-reconstruction-from-unconstrained-photo-collections,Skull2Face18,sanyal2019learning,wang2020lightweight,tewari2019FML,Thies3DV16} is limited. Thus, achieving high-fidelity face reconstruction from in-the-wild imagery usually relies on prior knowledge like 3DMMs~\cite{blanz1999morphable}, FLAME~\cite{li2017learning}, or DAM~\cite{lombardi2018deep}. For example, instead of directly regressing the geometry, MoFA~\cite{tewari2017mofa} uses a CNN-based image encoder to extract the semantically meaningful parameters (\eg, facial expression, shape, and skin reflectance) from a single 2D image and then uses the parametric model-based decoder to render the output image. Tran~\etal~\cite{tran2019learning} propose a weakly supervised model that jointly learns a nonlinear 3DMM and its fitting algorithm from 2D in-the-wild image collection. Gecer~\etal~\cite{gecer2019ganfit} train a facial texture generator in the UV space with self-supervision as their statistical parametric representation of the facial texture. Lin~\etal~\cite{lin2020towards} propose a GCN-based network to refine the texture generated by a 3DMM-based method with facial details from the input image. Yoon~\etal~\cite{yoon2019self} propose a network (I2ZNet) to learn a latent vector ${z}$ and head pose for DAM from a single image to reconstruct the texture and geometry.% The goal of this paper is to achieve high-fidelity face tracking from the in-the-wild imagery, thus we adopt the DAM~\cite{lombardi2018deep} decoder as our decoder.

\noindent \textbf{Lighting Estimation for In-the-Wild Imagery.}\quad
Most existing face reconstruction works~\cite{lin2020towards,9090444,sengupta2018sfsnet,tewari2018self,tewari2017mofa,towards-high-fidelity-nonlinear-3d-face-morphable-model,tran2019learning,Meka:2020} estimate the illumination using spherical harmonics (SH) basis function. For instance, Tewari~\etal~\cite{tewari2018self} regress the illumination parameters from the input image, and the rendering loss is computed after combining the estimated illumination and skin reflectance in a self-supervised fashion. In this way, it is hard to analyze the quality of the estimated illumination and skin albedo. Moreover, low-order spherical harmonics cannot produce hard shadows cast from point light sources, which will decrease the face reconstruction quality in many real-world scenarios. I2ZNet~\cite{yoon2019self} proposes a MOTC module to convert the color of the predicted texture to the in-the-wild texture of the input image. The MOTC can be viewed as a color correction matrix that corrects the white-balance between the two textures. However, the MOTC can only model the low-frequent lighting information, which will decrease the face registration performance if the lighting environment is complicated. Meanwhile, the surface information presented by a single image is limited, which leads to some artifacts such as over-smoothing and incorrect expression. To explicitly model the lighting, we propose a physics-based lighting model to learn the high-frequent lighting pattern (\eg, shading, and brightness) from data captured in a lab-controlled environment. Besides, a domain adaptation schema is proposed to bridge the domain mismatch between lab and wild environments.

\section{Adaptive Lighting Model}
\label{sec:method}
This section describes existing work on DAMs~\cite{lombardi2018deep} (Sec.~\ref{sec:dam}), construction of the lighting model with light-stage data
(Sec.~\ref{subsec:lighitng}), and the adaptation of the model for in-the-wild settings in Sec.~\ref{subsec:domaingap}.

\subsection{Deep Appearance Models}
\label{sec:dam}

Our work is based on the face representation described in DAM~\cite{lombardi2018deep}. It uses a variational auto-encoder (VAE)~\cite{Kingma2013} to jointly represent the 3D facial geometry and appearance that are captured from a multi-view capture light-stage. The decoder, $D$, can generate instances of a person's face by taking, as input, a latent code $\mathbf{z}$, which encodes the expression, and a vector $\mathbf{v}^v$ that represents viewing direction as a normalized vector pointing from the center of the head to the camera $v$:
\begin{equation}
\begin{aligned}
{\mathbf{\hat{M}}} , \mathbf{\hat{T}}^v\leftarrow \textit{D}(\mathbf{z}, \mathbf{v}^v) \enspace.
\end{aligned}
\label{eq:dam} 
\end{equation}
Here, $\mathbf{\hat{M}}$ denotes the 3D face mesh (geometry) and $\mathbf{\hat{T}}^v$, the view-dependent texture.
 
In this work, we assume the availability of a pre-trained DAM decoder of a subject, and propose a system to fit the model to images by estimating the rigid and non-rigid motion (i.e., facial expression). A major challenge in implementing such a system is how to account for illumination differences between the high controlled studio lighting system where the avatar was captured, and the in-the-wild captures where the avatar is deployed. 
 In the following, we will describe an adaptive lighting model that extends the original DAM formulation to enable high precision tracking in uncontrolled and complex environments. 

%The proposed face registration system reconstructs an avatar with realistic texture and accurate expression from video recorded under an in-the-wild environment (see Fig.XXXX). This section describes the architecture of the proposed approach. First, we introduce the critical component---lighting estimator in sec.~\ref{subsec:lighitng}. Then we explain how to eliminate the domain gap between fully-lit in light-stage and the in-the-wild lighting environment in Sec.~\ref{subsec:domaingap} for face registration.
%Fig.XX shows the overall diagram, which consists of four steps: Coarse Parameter Estimation (CPE), Coarse Lighting Estimation (CLE), Fine Parameter Estimation (FPE),

 \begin{figure}
\includegraphics[width= 1.0 \linewidth]{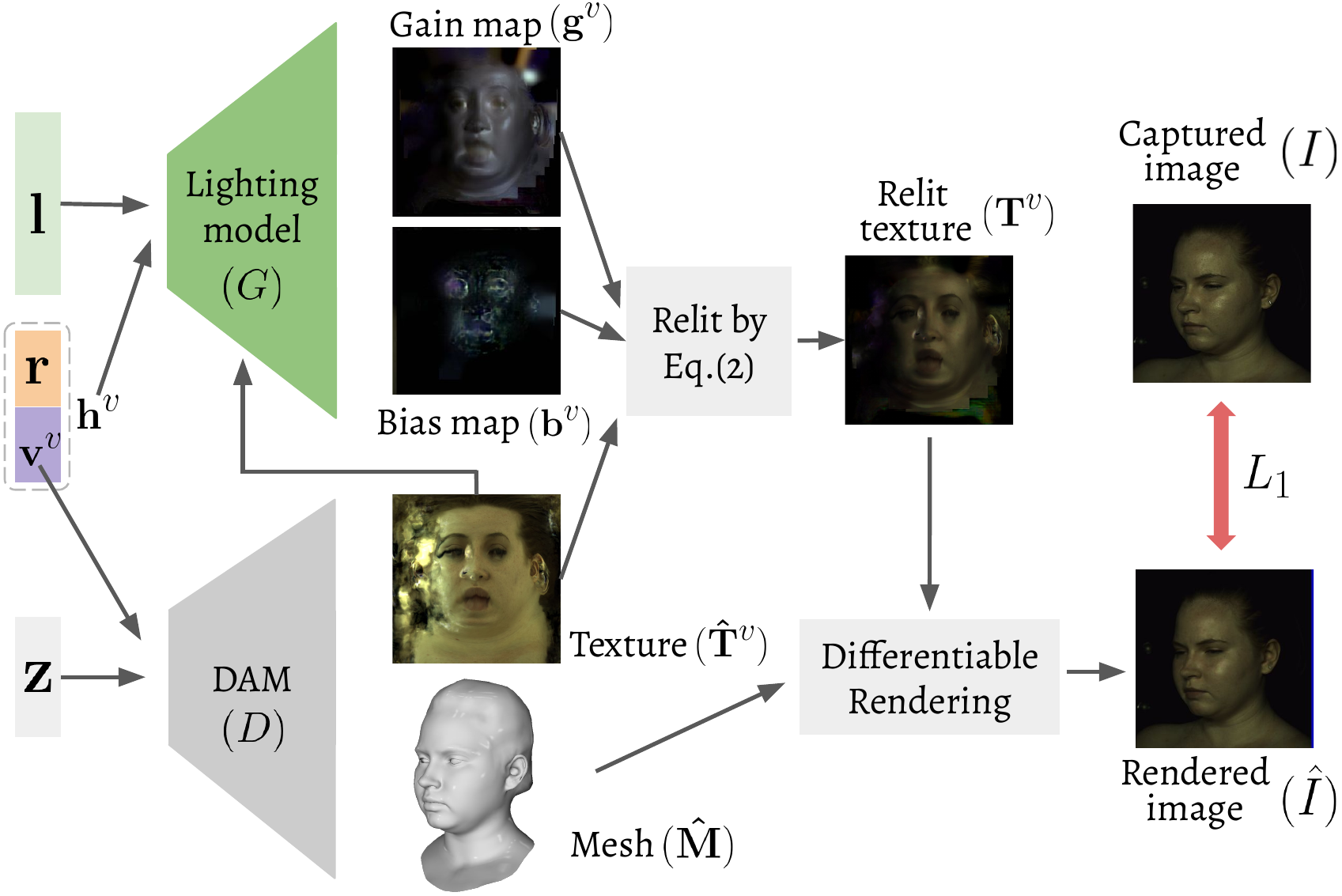}
\caption{Training the lighting model on the light-stage data. We update the lighting model $G$ and per-frame expression code $\mathbf{z}$ while fixing the other parameters.}
% %\vspace{-2mm}
\label{fig:mugsy_light_train}
\end{figure}

\subsection{Lighting Model}
\label{subsec:lighitng}
%At the core of our algorithm is a lighting model (see Fig.~\ref{fig:mugsy_light_train}), which aims at learning the illumination difference between fully-lit in light-stage and the target lighting in the in-the-wild image. 

In order to incorporate a generative model of lighting into the DAM formulation, we extend the capture system in~\cite{lombardi2018deep} to include 460 controllable lights that are synchronized with the multi-view camera system. The captured sequence was extended to include a portion where non-overlapping groups of approximately $10$ lights were turned on, interleaved with fully lit frames that were used for tracking. This data was used to build a relightable face model using the scheme illustrated in Figure~\ref{fig:mugsy_light_train}.

%With controllable lighting in a light-stage capture system, we are able to capture multiple views of a subject under different lighting configuration (including the fully-lit). This section, describes how to create the lighting model using these captured data, see Fig.~\ref{fig:mugsy_light_train}.

%We firstly track the coarse face mesh based on the multi-view images, to obtain the head rotation $\mathbf{r}$, translation $\mathbf{t}$ and the coarse face mesh ${\mathbf{M}}$ using the method from \cite{wu2018}. We then train a lighting model to transform the lighting from fully-lit DAM avatar to these captured data with different lighting patterns (Fig.~\ref{fig:mugsy_light_train}). 

Our formulation is inspired by the light-varying residual proposed by Nestmeyer~\etal~\cite{nestmeyer2020learning}, where illumination variations are represented using gain and bias maps, $\textbf{g}$ and $\textbf{b}$, each matching DAM's texture dimensions ($H\times W\times3$):
\begin{equation}
  \mathbf{T}^v = \mathbf{\hat{T}}^v \odot \mathbf{g}^v + \mathbf{b}^v,
  \label{eq:lighting_apply}  
\end{equation}
where $\odot$ denotes the element-wise product, $\mathbf{T}^v$ is the relit texture, and $\mathbf{\hat{T}}^v$ is the DAM avatar's original texture (fully-lit). The gain and bias maps depend on the lighting, head pose\footnote{Here, the rigid head pose consists of two parts: rigid rotation $\mathbf{r}\in \mathbb{R}^3$ and camera viewpoint vector $\mathbf{v}^v \in \mathbb{R}^3$. Similar to ~\cite{lombardi2018deep}, we assume that the viewpoint vector is relative to the rigid head orientation that is estimated from the tracking algorithm.}, viewpoint, and expression. These inputs, represented by $\mathbf{l}$, $\mathbf{h}^v$, and $\hat{\mathbf{T}}^v$, are processed by MLPs and are spatially repeated for concatenation with the DAM's texture followed by additional convolution operations to produce the final relit texture. This lighting model is, thus, defined as follows:
\begin{equation}
\begin{aligned}
{\textbf{g}^v} , \textbf{b}^v \leftarrow \textit{G}(\mathbf{l}, \mathbf{h}^v, \mathbf{\hat{T}}^v; \phi) \enspace,
\end{aligned}
\label{eq:lighting}  
\end{equation}
where $\phi$ denotes the weights of the network. Further details about $G$ are in Sec.~\ref{subsec:exp_setup} and in the supplementary material. 

Since the goal of our lighting model is to enable accurate registration in uncontrolled scenarios, we do not require a lighting representation that is geometrically interpretable, only one that can span the space of facial illuminations. As such, we represent lighting conditions using a one vector that specifies which of the lights, in each color channel, are active for a given training frame. We use a binary vector of 150 dimensions, which comprises the 50 lighting groups with three color channels each. The fully-lit frames are encoded as the all-one vector. In combination with the continuously parameterized head pose, this representation allows for continuous and smoothly varying illumination synthesis on the face that can model complex effects such as how shadows move as the
subject's head rotates in the scene. 

To train $G$, we take a pre-trained DAM decoder and fix its weights while minimizing the reconstruction error over all camera views in the subject's sequence, that is:
\begin{equation}
\begin{aligned}
\mathcal{L}_{\text{render}}(\phi, \mathbf{Z}) = \sum_{t,v} \left\lVert \left( I^v_t - \mathcal{R}(\mathbf{T}^v_t, \hat{\mathbf{M}}_t) \right) \odot m^v \right\rVert_1 \enspace,
\end{aligned}
\label{eq:mugsy_loss}  
\end{equation}
where the minimization is performed over the lighting model's weights, $\phi$, as well as the expression codes for each frame $\mathbf{Z} = \{\mathbf{z}_t\}$. In Equation~\ref{eq:mugsy_loss}, $m^v$ is a foreground mask from view $v$ and $\mathcal{R}$ is a differentiable rasterization function~\cite{parker2010optix}. To ensure stable convergence, we employ $L_2$-shrinkage on the expression codes, $\mathbf{Z}$, and use the technique in~\cite{wu2018} to obtain a tracked mesh, $\mathbf{M}_t$, from the fully-lit frames, that are used to geometrically constrain the optimization, resulting in the following total objective:
\begin{equation}
\mathcal{L} = \mathcal{L}_{\text{render}} + \lambda_{\text{geo}} \sum_t \| \mathbf{M}_t - \hat{\mathbf{M}}_t \|^2 + 
\lambda_{\text{reg}} \| \mathbf{Z} \|^2.
\end{equation}
Regularization weights of $\lambda_{\text{geo}}=1.0$ and $\lambda_{\text{reg}} = 0.1$ were chosen for all experiments in Sec.~\ref{sec:exp} by cross validation.

\begin{algorithm}[t]
\caption{Lighting Model Adaptation}
  \begin{algorithmic}
    \State {\bf Input:} lighting model $G$ with weights $\phi$, $K$ key frames with initial face parameters $\{(I,\mathbf{\tilde{p}})_k\}$, camera viewpoint vector $\mathbf{v}^v$
    \State {\bf Output:} adapted lighting model $G$ and lighting code $\mathbf{l}$
    \State {\bf Initialization:} set $\mathbf{l}$ to zeros, $\phi$ to pre-trained weights by Sec.~\ref{subsec:lighitng}, face parameters $\{\mathbf{p}_k\}$ to $\{\mathbf{\tilde{p}}_k\}$
    \For {number of iterations}
      \State \textbf{\# Fitting} $\mathbf{l}$
      \For {number of iterations}
        \State Unfreeze $\mathbf{l}$, freeze $\phi$ and $\{\mathbf{p}_k\}$
        \State Calculate $\mathcal{L}_{pix}$ using Eq.~\ref{eq:pix_loss}
        \State $\mathbf{l} \leftarrow Adam\{ \mathcal{L}_{pix}\}$
      \EndFor
      \State \textbf{\# Fitting} $\phi$ and $\{\mathbf{p}_k\}$
      \For {number of iterations}
        \State Freeze $\mathbf{l}$, unfreeze $\phi$ and $\{\mathbf{p}_k\}$
        \State Calculate $\mathcal{L}_{pix}$ using Eq.~\ref{eq:pix_loss}
        \State $\phi, \{\mathbf{p}_k\} \leftarrow Adam\{ \mathcal{L}_{pix}\}$
      \EndFor
    \EndFor
  \end{algorithmic}
\label{al:al1}

\end{algorithm}

\begin{figure*}[t]\centering
\includegraphics[width= 0.9\linewidth]{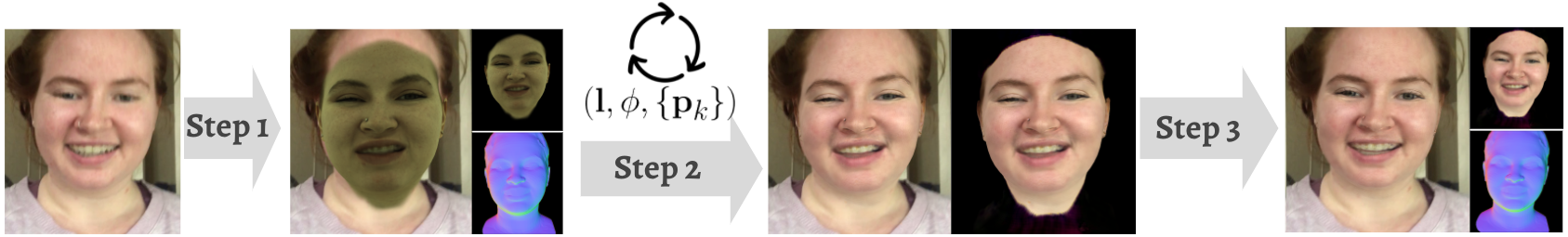}
\caption{The pipeline of in-the-wild registration. We estimate the initial tracking parameters in \textbf{step 1}, adapt the lighting model and tracking parameters $\mathbf{l}, \phi,\{\mathbf{p}_k\}$ with $K$ reference frames in \textbf{Step 2}, and further optimize the tracking parameters in \textbf{step 3}.}
% %\vspace{-2mm}
\label{fig:in-the-wild}
\end{figure*}

 \begin{figure}[t]\centering
\includegraphics[width= 0.95\linewidth]{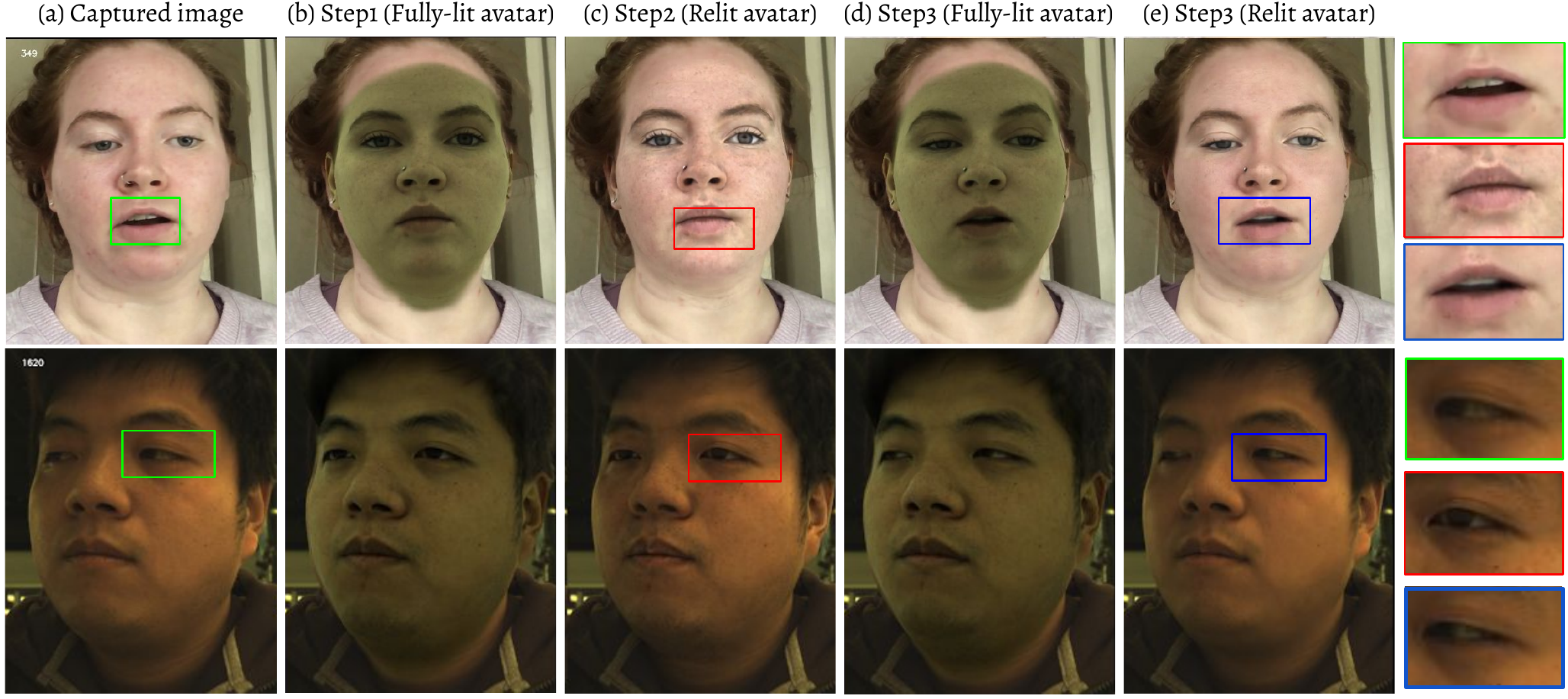}
\caption{Result of each step in Sec.~\ref{subsec:domaingap}. From left to right: (a) captured image, (b) avatar output by \textbf{step 1}, (c) relit avatar after \textbf{step 2}, (d) fully-lit avatar after step 3, and (e) relit avatar after \textbf{step 3}. The last column shows the zoom-in of corresponding color rectangle, please notice the lip shapes and gaze directions. }
\vspace{-3mm}
\label{fig:in_the_wild_results}
\end{figure}

\subsection{Registration In-the-Wild}
\label{subsec:domaingap}

With the lighting model described in Sec.~\ref{subsec:lighitng}, we have a personalized face model that can synthesize realistic variations in expression as well as lighting suitable for an analysis-by-synthesis approach to register the model to in-the-wild video with uncontrolled illumination. However, to achieve robust and precise results, special care needs to be applied in how registration is performed. Our algorithm comprises three steps (Fig.~\ref{fig:in-the-wild}) outlined below, each designed to address initialization, accuracy, and computational efficiency, respectively. 

\noindent\textbf{Step 1: Initialization.}\quad To avoid registration terminating in poor local minima, we initialize the pose and expression parameters by matching against facial keypoints in the image found via an off-the-shelf detector (\eg,\cite{lee2015face}). Specifically, face landmarks, $\{{L}_i\}$, describing facial features such as eye corners, lip contours, and the face silhouette correspond to fixed vertices in the face model's geometry, $\{\ell_i\}$. The initial face parameters, $\mathbf{\tilde{p}} = [\mathbf{\tilde{r}}, \mathbf{\tilde{t}}, \mathbf{\tilde{z}}]$, are then found by minimizing the reprojection error over these landmark points in all camera views, $v$, for every frame:
\begin{equation}
  \mathcal{L}_{\text{land}}(\mathbf{\tilde{p}}) = \sum_{v,i} \left\lVert \Pi_v \left( \mathbf{\tilde{r}} \mathbf{\tilde{M}}^{(\ell_i)} + \mathbf{\tilde{t}} \right) - L_i^v \right\rVert ^2,
% %\vspace{-1mm}
\end{equation}
where $\Pi_v$ is the projection operator based on camera parameters that are assumed to be available. The face mesh, $\mathbf{\tilde{M}}$, is calculated using Eq.~\ref{eq:dam} with the expression code $\mathbf{\tilde{z}}$. Due to the landmarks' sparsity and detection noise, minimizing $\mathcal{L}_{\text{land}}$ results in only a rough alignment of the face in each frame (\eg, Fig.~\ref{fig:in_the_wild_results} (b)). Nonetheless, it places the model within the vicinity of the solution, allowing the more elaborated optimization procedure described next to converge. 

\noindent\textbf{Step 2: Lighting Model Adaptation.}\quad Although the lighting model described in Sec.~\ref{subsec:lighitng} equips us with the ability to synthesize variations in facial illumination, using the light-stage to simulate the total span of lighting variations encounter in the wild remains challenging. Effects such as nearfield lighting, cast shadows, and reflections from nearby objects are commonly observed in uncontrolled settings. To account for variations not spanned by our lighting model, $G$, in addition to solving for the lighting parameters, $\mathbf{l}$, we simultaneously fine-tune the model's weights, $\phi$, to obtain a better fit to in-the-wild images. Specifically, we minimize the following loss over a collection of $K$ frames~\footnote{We can select K frames out of the testing sequence for adaptation if the lighting of the K frames is the same as the testing sequence.} from the target environment:
\begin{equation}
\begin{aligned}
\mathcal{L}_{\text{pix}}(\mathbf{l}, \phi, \{\mathbf{p}_k\}) &= \sum_k \left \lVert {r}_k \odot w_k \right\rVert_1 + \lambda_{\triangle} \left\lVert \triangle {r}_k \odot w_k \right\rVert_1  \enspace,
\end{aligned}
\label{eq:pix_loss}  
\end{equation}
where ${r}_k=I_k - \hat{I}(\mathbf{p}_k)$ is the reconstruction residual, $w_k$ is the foreground mask, and $\mathbf{p}_k = [\mathbf{r}_k, \mathbf{t}_k, \mathbf{z}_k]$. Here, $\Delta$ denotes the image Laplacian operator that makes the loss more robust to residual differences due to illumination and generally improves results.

\noindent\textbf{Step 3: Face Tracking.}\quad The procedure described previously can generate accurate estimates of facial expression, but requires batch processing to adapt the lighting model simultaneously over several frames. However, once the lighting model has been adapted, the parameters $\mathbf{p} = [\mathbf{r}, \mathbf{t}, \mathbf{z}]$ for any new frames can be estimated independently of the lighting model $G$. Thus, in practice, we adapt the lighting model using only a small subset of $K$ frames and estimate $\mathbf{p}$ with the updated model $G$. To further improve accuracy, in addition to the optimizing the loss in Eq.~\ref{eq:pix_loss}, similarly to \cite{cao2018stabilized}, we use dense optical flow~\cite{Kroeger2016} between the rendered model and the image to further constrain the optimization. It is computed over all projected mesh vertices in all camera views as follows:
\begin{equation}
\begin{aligned}
  \mathcal{L}_{\text{flow}}(\mathbf{p}) = \sum_{v,i} \left\lVert \left( \mathbf{r} \mathbf{M}^{(i)} + \mathbf{t} \right) - \Pi_v \left( \mathbf{\tilde{r}} \mathbf{\tilde{M}}^{(i)} + \mathbf{\tilde{t}} \right) - \mathbf{d}_i^v\right\rVert ^2,
\end{aligned}
\end{equation}
where mesh $\mathbf{M}$ and $\mathbf{\tilde{M}}$ are calculated using Eq.~\ref{eq:dam} with the latent face code $\mathbf{z}$ and $\tilde{\mathbf{z}}$ respectively, and $(\tilde{\mathbf{r}}, \tilde{\mathbf{t}}, \tilde{\mathbf{z}})$ are initial parameters from \textbf{Step 1}. We optimize the per-frame face parameters $\mathbf{p}=\{\mathbf{r}, \mathbf{t}, \mathbf{z}\}$ by minimizing the total loss $\mathcal{L} = \mathcal{L}_{\text{pix}} + \lambda_{\text{flow}}\mathcal{L}_{\text{flow}}$, where $\lambda_{\text{flow}} = 3.$0 was chosen via cross validation. Fig.~\ref{fig:in_the_wild_results} (d-e) shows some examples of results obtained through this process, demonstrating accurate alignment and reconstruction of lip shape and gaze direction that were absent in \textbf{Step 1}.

 \begin{figure*}
 \centering
\includegraphics[width= 0.95\linewidth]{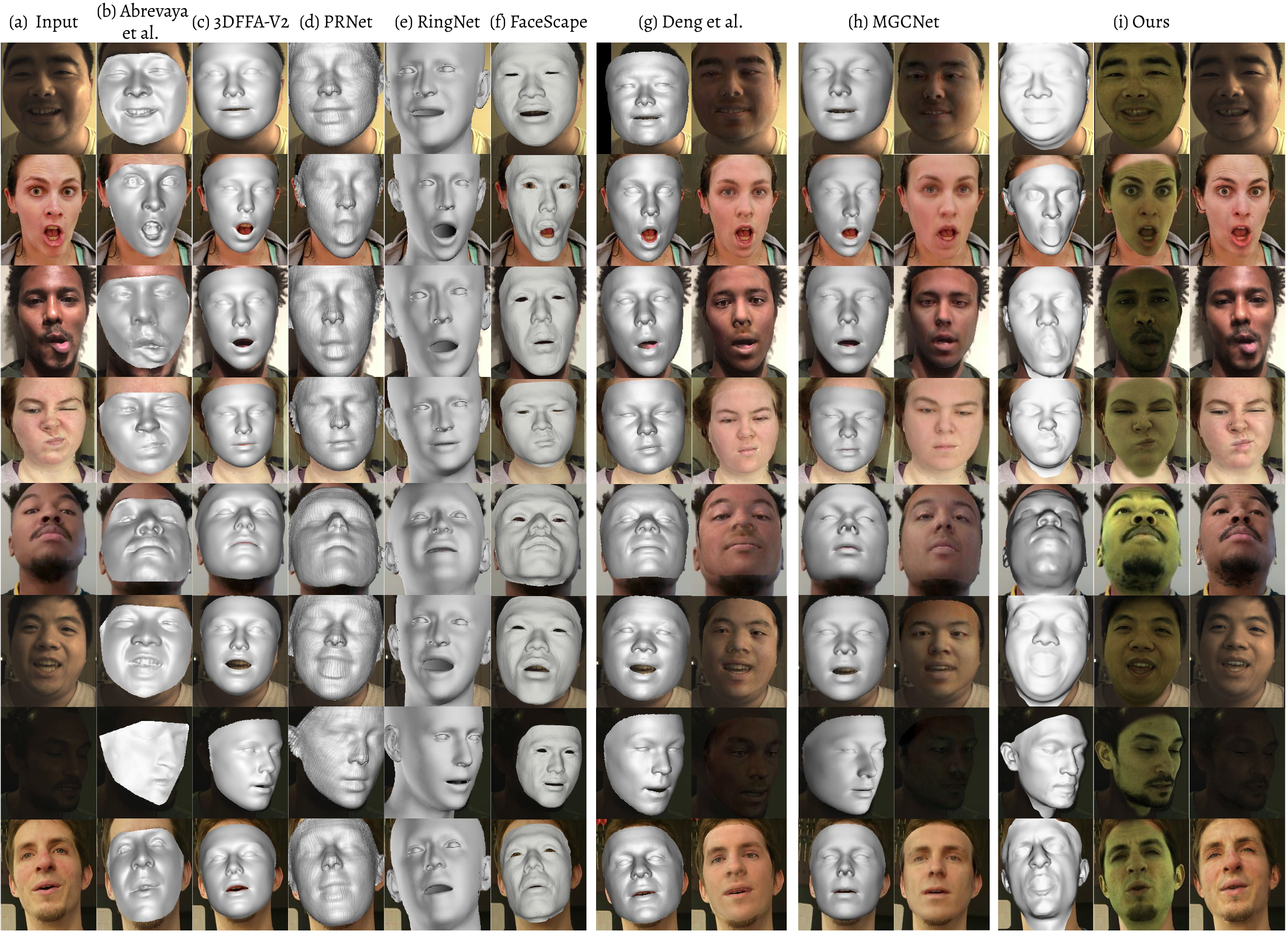}

\caption{Qualitative comparison. We suggest to view it using a monitor for better visual quality. We selected $8$ subjects from the test set with different lighting conditions, facial expression, and head motion. From left to right: (a) Captured image, (b) Abrevaya~\etal~\cite{abrevaya2020cross}, (c) $\text{3DDFA}{v2}$~\cite{guo2020towards}, (d) PRNet~\cite{feng2018joint}, (e) RingNet~\cite{sanyal2019learning}, (f) FaceScape~\cite{yang2020facescape}, (g) Deng~\etal~\cite{deng2019accurate}, (h) MGCNet~\cite{shang2020self}, and (i) our method.}
\vspace{-2mm}
\label{fig:mesh_res}
\end{figure*}

 \begin{figure}[t]\centering
\includegraphics[width= 0.95 \linewidth]{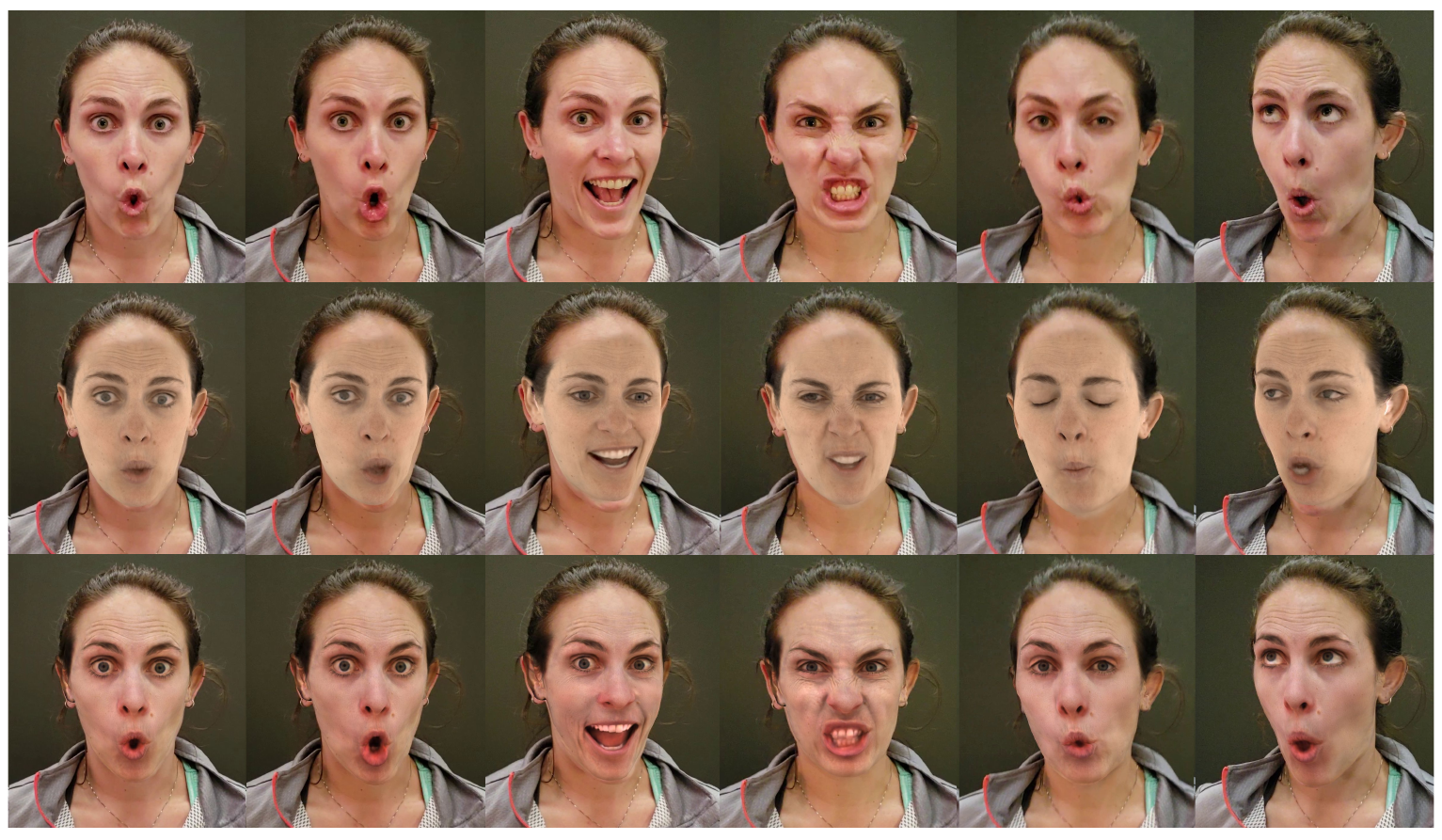}
\caption{Visual comparison between our method and I2ZNet~\cite{yoon2019self} on testing video frames. From top to bottom: captured image, I2ZNet~\cite{yoon2019self}, and our method.}
\vspace{-5mm}
\label{fig:compare_i2z}
\end{figure}

 \begin{figure*}[t]
\centering
\includegraphics[width= 0.98\linewidth]{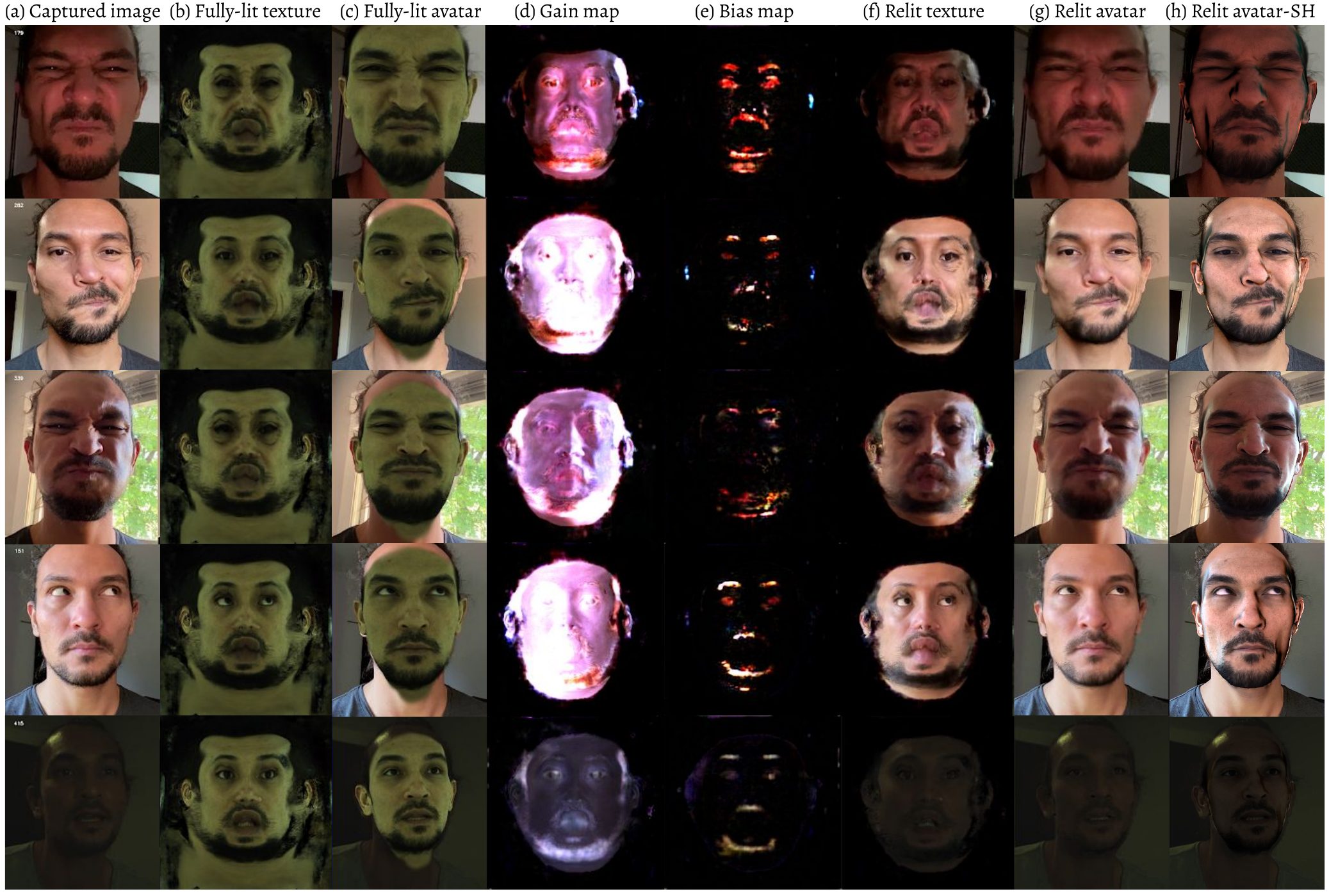}
\caption{Visual results under different lighting conditions. Besides our final relit avtar (g), we also show the tracked avatar (h) using spherical harmonics (SH) as illumination model. Our method can handle different lighting conditions well. }
\vspace*{-2mm}
\label{fig:texture_compare}
\end{figure*}

\section{Experiments}
\label{sec:exp}
% %\vspace*{-1mm}
In this section, we conduct quantitative and qualitative experiments to show the performance of the proposed face tracking framework on {\em in-the-wild} videos. Sec.~\ref{subsec:exp_setup} explains dataset and implementation details. Sec.~\ref{subsec:comparison} compares our method to state-of-the-art methods, and Sec.~\ref{subsec:ablation} describes the ablation studies.

\subsection{Experimental Settings} 
\label{subsec:exp_setup}
%\vspace*{-1mm}

\noindent \textbf{Dataset Collection.}\quad
%We record two different kinds of video data: light-stage data and in-the-wild data. Both of them comprise a diverse set of facial expressions captured under various lighting conditions.
We recorded our light-stage data in a calibrated multi-view light-stage consisting of $40$ machine vision cameras capable of synchronously capturing HDR images at $1334 \times 2048$ / $90$ fps and a total of $460$ white LED lights. We flash a group of LEDs (at most $10$) per frame and instruct our subjects to make diverse expressions with head movements. There are $50$ different lighting patterns and one fully-lit pattern. We record a total of $13$ minutes video sequence of one subject (see supplementary videos). 
%During training (Sec.~\ref{subsec:lighitng}), we only capture train our lighting model on one subject and use other three subjects as validation set since we found that our lighting transfer network is not sensitive to identities and can be easily adapted to new identity be $\text{IE}_{Gl}$ described in Sec.~\ref{subsec:domaingap}. %Among the $40$ camera views, we select $10$ views that cover different view directions.
%Our goal is to apply the lighting module on in-the-wild videos for high-fidelity face registration. Thus, To demonstrate the effectiveness of our lighting model and face tracking system, we also record the in-the-wild data using commercial cameras with diverse lightings, poses and facial expressions from different identities as testing video data. We collect videos from $10$ subjects, with around $5$ video clips under various lighting conditions for each subject. Those videos are either recorded by iPhone or a binocular camera. 

The in-the-wild video test were gathered using the frontal camera of an iPhone. We captured videos for $10$ subjects. We collected around $5$ video clips for each subject, performing different facial expressions and head movements, under various lighting conditions and environments.

 \noindent \textbf{Implementation Details.} \quad The light-stage training step (Sec.~\ref{subsec:lighitng}) and the lighting adaptation step cost 36 hours and 4mins, respectively, on an NVIDIA DGX machine. In all our experiments, we used the Adam optimizer~\cite{kingma2014adam} to optimize the losses. In order to cover more lighting space during the training of the lighting transfer module, we augmented the RGB channels of the light-stage lighting color with randomly selected scales, which are also used to scale the lighting code $\mathbf{l}$ in the training data. 
 The architecture of our lighting decoder $G$ is as follows: we first encode the input head pose $\mathbf{h}^v$ and $\mathbf{l}$ with two MLPs to $256$ dimensions. After concatenating the two latent features, we pass it to a fully-connected layer and a convolution layer followed by four transposed convolutions with each layer. The fully-lit texture $\hat{\mathbf{T}}^{v}$ is encoded by four convolution layers with each layer followed by a down-sampling layer to texture feature. Then we concatenate the texture feature and lighting feature and pass it to two separate branches consists of two transpose convolution. The two branches output gain map $\textbf{g}_t^v$ and bias map $\textbf{b}_t^v$ with the resolution of $256 \times 256$, and we upsample them $4$ times using bilinear
interpolation to the same resolution as texture $\hat{\mathbf{T}}^{v}$. Please refer to supplementary materials for details. While training the lighting model on the light-stage data in Sec.~\ref{subsec:lighitng}, the rendering loss is optimized with an initial learning rate of $1e^{-3}$, which is decreased by a quarter after every $10$ epochs.

During the registration of the {\em in-the-wild} videos in Sec.~\ref{subsec:domaingap}, we fit the DAM code $\mathbf{z}_k$ in \textbf{step 1} with $1000$ iterations. The initial learning rate is $0.1$, and we decrease it by half after every 500 iterations. In \textbf{step 2}, the $K$ frames are uniformly sampled according to the value of $\tilde{\mathbf{r}}$ to cover diverse head movements. We fit the lighting code $\mathbf{l}$ with $250$ iterations with a learning rate of $1e^{-2}$, then update $G$'s network parameter $\phi$ and face parameters $\{\mathbf{p}_k\}$ with $500$ iterations with a learning rate of $1e^{-3}$. We alternatively update $\mathbf{l}$, $\phi$, and $\{\mathbf{p}_k\}$ for $4$ times. In \textbf{step 3}, we update the face parameters $\{\mathbf{p}_k\}$ with the same hyper-parameters as \textbf{step 1}.

\noindent \textbf{Evaluation Metrics.} \quad We used a variety of perceptual measures to quantitatively compare the registered image against the ground-truth in-the-wild image. Besides the pixel-level $L_2$ distance, we adopted PSNR and structural similarity (SSIM)~\cite{wang2004image} for human perceptual response. To evaluate the realism of the output avatar, we computed the cosine similarity (CSIM) between embedding vectors of the state-of-the-art face recognition network~\cite{deng2019arcface} suggested by~\cite{zakharov2019few,chen2020talking} for measuring identity mismatch between the input image and reconstructed avatar.

\subsection{Comparison with state-of-the-art methods}
\label{subsec:comparison}
%\vspace*{-1mm}
To demonstrate the effectiveness of our lighting model and the face tracking quality, we compared our algorithm against the following set of related methods using the pretrained models provided by the authors: Abrevaya~\etal~\cite{abrevaya2020cross}, $\text{3DDFA}{v2}$~\cite{guo2020towards}, PRNet~\cite{feng2018joint}, RingNet~\cite{sanyal2019learning}, FaceScape~\cite{yang2020facescape}, Deng~\etal~\cite{deng2019accurate}, MGCNet~\cite{shang2020self}, and I2ZNet~\cite{yoon2019self}. As the baseline method, we used the same face registration method proposed in Sec.~\ref{sec:method}, but use the standard spherical harmonics (SH)~\cite{muller2006spherical} illumination model, which is denoted as ours-SH. We adopt the same parameter setting as~\cite{tewari2017mofa} for SH, and train a regression network to regress the input images to the $27$ dimensional SH parameters.

\noindent \textbf{Qualitative Evaluation.} \quad
Fig.~\ref{fig:mesh_res} shows the tracked geometry results for different methods~\footnote{Note that our method relies on person-specific DAM.}. We can observe that our proposed method is robust to diverse facial expressions, poses, and lighting conditions. For example, in the $3^{rd}$ and $4^{th}$ row, all other methods failed to describe the expression (\eg, lips) except our method and Abrevaya~\etal~\cite{abrevaya2020cross}. In the $7^{th}$ row, all other methods can not output the correct head pose, while our method can still reconstruct high-quality geometry and texture under dark lighting conditions. We also show the reconstructed avatars by (g)~\cite{deng2019accurate}, (h)~\cite{shang2020self}, and (i) our method (fully-lit and relit avatar). Comparing with other methods, our method not only generates a more realistic avatar, but also considers the lighting details (\eg, shadows and specular highlights, see the relit avatar in $1^{th}$, $3^{rd}$, and $6^{th}$ row). Furthermore, we compare our relit avatar with I2ZNet~\cite{yoon2019self} in Fig.~\ref{fig:compare_i2z}. Although I2ZNet is a person-specific model, our method produces better visual results. 

Fig.~\ref{fig:texture_compare} shows the visual results under different real-world lighting environments. The proposed method is robust to different unseen lighting conditions, and our face tracking system can output a high-quality avatar with the aid of our lighting model. Fig.~\ref{fig:texture_compare}(h) shows the tracking results using the SH illumination model. The reconstruction error between the captured frame and the avatar relit by SH model is large, and it decreases the face tracking performance.

\begin{figure}[t]
\includegraphics[width= 1.0 \linewidth]{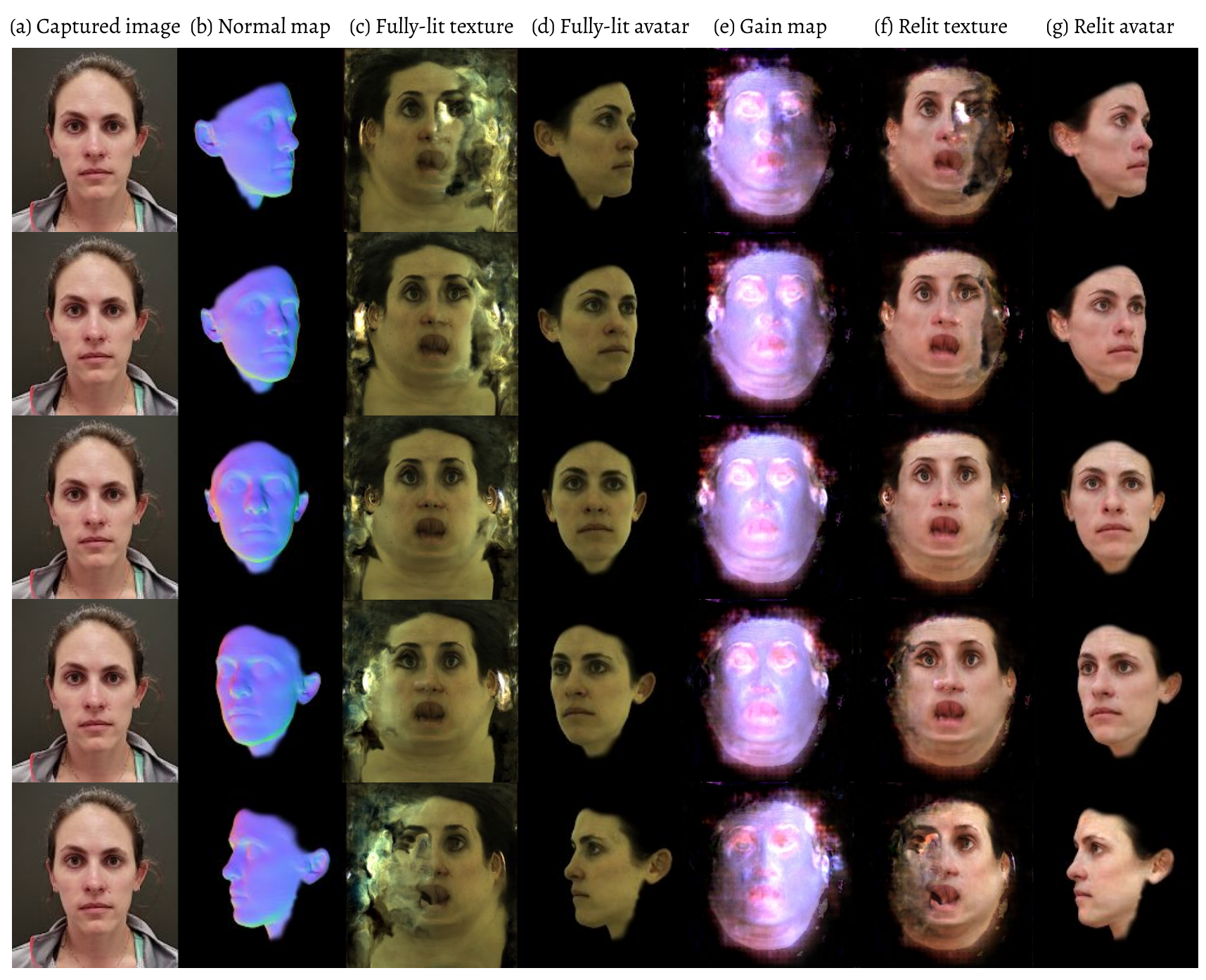}
\caption{The visual results for different viewpoints. From left to right: captured image, normal map, fully-lit texture, fully-lit avatar, gain map, relit texture, and relit avatar. From top to bottom: different viewpoints. Please notice the changes of the gain map due to different viewpoints.}
% %\vspace{-2mm}
\label{fig:ablation_rotate}
\end{figure}

We also show our avatar rendered from different viewpoints in Fig.~\ref{fig:teaser} and Fig.~\ref{fig:ablation_rotate}. We can find that our method can output the high-fidelity avatar from any viewpoint. Our lighting model is conditioned on the camera viewpoints, so the gain map and bias map will be adjusted to match the lighting in the specific view. Please refer to the supplementary video for more visual results.

\noindent \textbf{Quantitative Evaluation.} \quad
 Tab.~\ref{tab:quantative_tb} shows the quantitative results of MGCNet~\cite{shang2020self}, Deng \etal~\cite{deng2019accurate}, I2ZNet~\cite{yoon2019self}, and our method. Our method outperforms other methods not only at the human perception level, but also at the identity preserving ability (CSIM score). The high CSIM score indicates that our reconstructed avatar produces high identity similarity. We can see that the $L_2$ and other perceptual scores of the avatar optimized with the SH illumination model result in higher error, although it preserves the identity information (CSIM). 

%%%%%%%%%%%%%%%%%%%%%%%%%%
\begin{table}[t]
% \footnotesize
 \centering
\begin{tabular*}{0.95\linewidth}{ l c c c c }
   \toprule
   \hline
  %  \toprule
Methods & \multicolumn{4}{c}{Metrics} \\ \hline
  %  \midrule 
  % \hline
& {$L_2$ $\downarrow$} & {SSIM$\uparrow$}& {PSNR$\uparrow$} & {CSIM$\uparrow$} \\
\hline
 {MGCNet~\cite{shang2020self}} &24.56 & 0.86 & 34.47 & 0.305  \\ %\hline

 {Deng~\etal~\cite{deng2019accurate}} & 27.33  &0.86  & 33.81 & 0.434 \\ 
 {I2ZNet~\cite{yoon2019self}} & 20.49  & 0.927 &35.03 & 0.592 \\  
 %\hline
 {Ours-SH} & 41.01 & 0.79 & 32.00 & 0.693 \\
 { Ours} & \textbf{12.59} &\textbf{0.93} &\textbf{ 37.93} &\textbf{0.871}\\ %\hline
   \bottomrule
 \end{tabular*}
 \caption{The quantitative evaluation on the test set. $\downarrow$/$\uparrow$ denote the lower/higher, the better. The top-1 scores are highlighted. }
 \label{tab:quantative_tb}
 \vspace{-4mm}
\end{table}
%%%%%%%%%%%%%%%%%%%%%%%%%

%%%%%%%%%%%%%%%%%%%%%%%%%%
\subsection{Ablation Study}
\label{subsec:ablation}
% %\vspace*{-1mm}
We have already shown the superiority of our face tracking algorithm along with the lighting model over other methods. To further demonstrate the effectiveness of different steps in Sec.~\ref{sec:method}, we make a comprehensive ablation study on a subset of the testing videos.

\begin{figure}[t]
\includegraphics[width= 0.98 \linewidth]{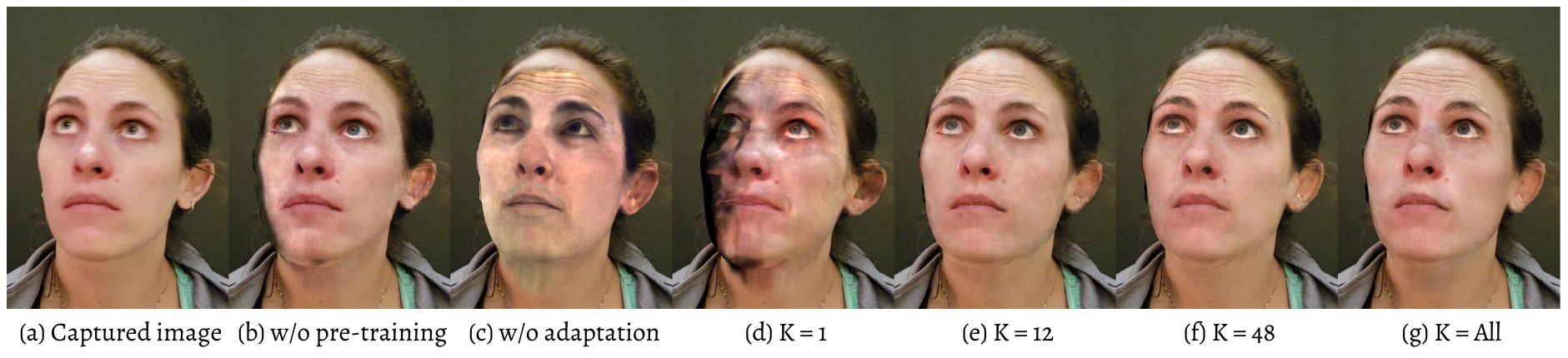}
\caption{Visual comparison of the ablation study.}
\vspace{-4mm}
\label{fig:ablation}
\end{figure}

\noindent \textbf{The lighting model.}\quad
To evaluate the role of our lighting model, we test our face tracking system with the lighting model under two different settings: without pre-training on light-stage data (w/o Pre-train.) and without lighting adaptation on wild video frames (w/o Adapt.). We set $K=48$ in the without pre-training experiment. Tab.~\ref{tab:ablation_tb} shows the quantitative evaluation results, and we can see that all the scores are improved with light-stage pre-training ($1^{st}$ and $5^{th}$ row). Although without adaptation ($2^{nd}$ row) performs slightly better than $K=1$, with more reference frames for adaptation, the adapted lighting model performs much better than without adaptation. Fig.~\ref{fig:ablation} shows the visual comparison. We can see that the pre-training on light-stage data and lighting adaptation both contribute to the final tracking results, where the pre-training on light-stage data enables the lighting interpretation ability, and the adaptation step enables the lighting model to generate accurate gain and bias map for target video frames.

\noindent \textbf{Influence of the value of $K$.}
\quad 
To evaluate the effect on the number of reference frames used in the lighting model adaptation step, we sample reference frames with different $K$ and keep other settings the same, and show the results in Tab.~\ref{tab:ablation_tb} and Fig.~\ref{fig:ablation}. We can find that with only 48 reference frames, the lighting model can be perfectly adapted to the target video and achieve comparable results in visual metrics ($L_2$, SSIM, and PSNR). If the amount of the selected reference frames is too small (\eg, $K=1$), the lighting model will be over-fitted to the selected frames and loss the lighting interpolation ability.
% %\vspace*{-1mm}

\begin{table}[t]
% \footnotesize
 \centering
\begin{tabular*}{0.95\linewidth}{ l c c c c }
   \toprule
   \hline
  %  \toprule
Methods & \multicolumn{4}{c}{Metrics} \\ \hline
  %  \midrule 
  % \hline
& {$L_2$ $\downarrow$} & {SSIM$\uparrow$}& {PSNR$\uparrow$} & {CSIM$\uparrow$} \\
\hline
% w/o Pretraining & 30.25  & 0.84 & 33.47  & 0.692 \\ 
w/o Pre-train. & 14.88  & 0.941 & 37.08 & 0.783  \\ 
w/o Adapt. & 18.03 & 0.938 & 35.59  & 0.593  \\
%\hline
% w/o Perceptual loss &  &  &  &  \\  %\hline
% {Ours-SH} &  &  &  & \\  
$K$ =1 & 18.58 & 0.900 & 35.48 & 0.550  \\  %\hline
$K$ =12 & 12.30  & 0.950 & 37.32 &0.842  \\  %\hline
{$K$ = 48} & {11.35} &{0.950} &{37.66 } &{0.809 }\\ %\hline
$K$ = All & 11.33  & 0.952 & 37.78 & 0.863  \\  %\hline
% Ours-12-part & 12.32  & 0.950 & 37.31 & 0.796  \\  %\hline
% Ours-48-part & 11.88 & 0.950 & 37.46 & 0.813  \\  %\hline
% Ours-All-part & 12.52  & 0.942 & 37.31 & 0.799 \\  %\hline
   \bottomrule
 \end{tabular*}
 \caption{Quantitative results of the ablation study.}
 \label{tab:ablation_tb}
%\vspace{-3mm}
\end{table}

\section{Conclusion}
%\vspace*{-1mm}
We present a new in-the-wild face tracking algorithm with an adaptive lighting model that can infer a high-fidelity 3D avatar. The proposed lighting model inherits the prior knowledge from the light-stage data, and it adapts to in-the-wild samples to produce high-quality re-lighting results for our face tracking. Results confirm that relatively few adaption samples ($48$) are enough to produce hyper-realistic results in the avatar. While the proposed method can generate photo-realistic avatars from videos in the wild, our lighting model assumes that the lighting source in the testing video is fixed, and it uses one single lighting code to represent the lighting in the whole video sequence. This is a limitation of the current model, and the model can produce undesirable results if the lighting is changing in the video. In the future, we will explore on-line adaptation methods to address this limitation of the current work.

\noindent \textbf{Acknowledgement.} \quad This work was done when Lele Chen was an intern at Facebook Reality Labs. 
%This work was partially supported by NSF IIS 1741472, IIS 1813709. %This article solely reflects the opinions and conclusions of its authors and not the funding agents.

{\small
\bibliographystyle{ieee}
\bibliography{ref}
}

\clearpage

\appendix
\begin{center}
\textbf{\large Supplemental Materials}
\end{center}

\begin{figure}[t]
\includegraphics[width=0.98\textwidth]{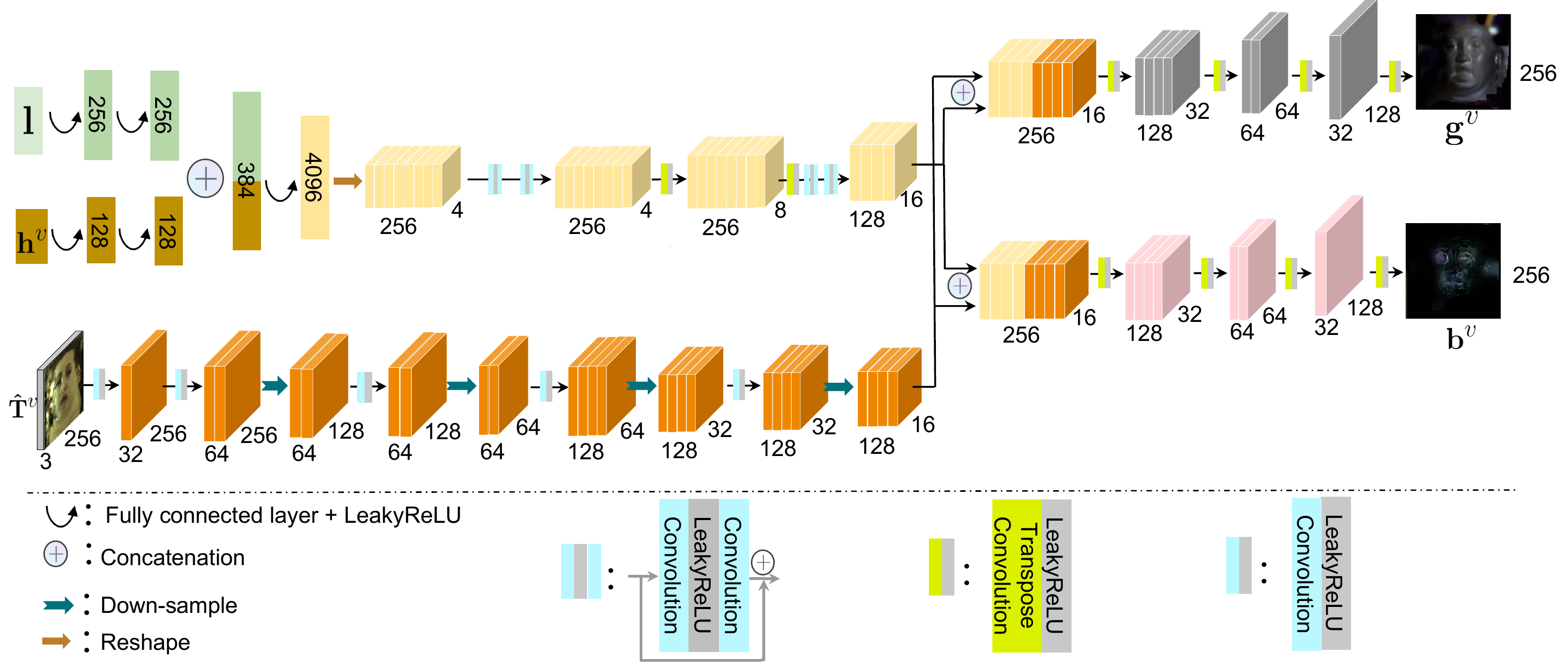}
\captionof{figure}{The detailed network structure of our lighting transfer network ($G$).}
\label{fig:teaser2}
\end{figure}

\section{Network Structure}
\label{subsec:net_str}
We present the detailed network structure in Fig.~\ref{fig:teaser2}. 

\section{Inputs and Outputs}
\label{subsec:input}
The lighting code $\mathbf{l}$ is a pre-defined vector when we train $G$ on light-stage data, and is a learnable vector when we refine $G$ on in-the-wild video frames. During training on light-stage data, the lighting direction is encoded by the position of the non-zero element in $\mathbf{l}$, and the lighting color is encoded by the value of the non-zero element in $\mathbf{l}$. The view-dependent head pose $\mathbf{h}^v \in \mathbb{R}^6 = \{  \mathbf{r}_x, \mathbf{r}_y, \mathbf{r}_z, \mathbf{v}^v_x,\mathbf{v}^v_y,\mathbf{v}^v_z \}$, where $\mathbf{r} = \{\mathbf{r}_x, \mathbf{r}_y, \mathbf{r}_z\}$ and $\mathbf{v}^v=\{\mathbf{v}^v_x,\mathbf{v}^v_y,\mathbf{v}^v_z\}$ are rigid head rotation and viewpoint vector, respectively. The fully-lit texture $\hat{\mathbf{T}}^v$ is obtained from DAM decoder, and we down-sample it to the size of $3\times256\times256$. 

The outputs are the gain and bias map $\mathbf{g}^v, \mathbf{b}^v$, and we upsample the output $\mathbf{g}^v, \mathbf{b}^v$ back to the size of $3\times1024\times1024$ by bilinear interpolation.

\end{document}